\documentclass[lettersize,journal]{IEEEtran}
\usepackage{amsmath,amsfonts}
\usepackage{algorithmic}
\usepackage{algorithm}
\usepackage{array}
\usepackage[caption=false,font=normalsize,labelfont=sf,textfont=sf]{subfig}
\usepackage{textcomp}
\usepackage{url}
\usepackage{verbatim}
\usepackage{graphicx}
\usepackage{cite}
\usepackage{multirow}
\usepackage{xspace}
\usepackage[dvipsnames, svgnames]{xcolor}
\usepackage{enumitem}
\usepackage{wrapfig}
\usepackage[normalem]{ulem}
\usepackage{comment}
\usepackage{amssymb}
\usepackage{amsmath}
\usepackage{mathtools}

\usepackage{dblfloatfix}    

\usepackage{array}
\newcolumntype{L}[1]{>{\raggedright\let\newline\\\arraybackslash\hspace{0pt}}m{#1}}
\newcolumntype{C}[1]{>{\centering\let\newline\\\arraybackslash\hspace{0pt}}m{#1}}
\newcolumntype{R}[1]{>{\raggedleft\let\newline\\\arraybackslash\hspace{0pt}}m{#1}}
\newcolumntype{P}[1]{>{\raggedright}p{#1}}

\newcommand\myparagraph[1]{ \vspace{4pt} \noindent \textbf{#1.}}

\newcommand{\system}{\textsc{Mountaineer}\xspace}

\usepackage{tcolorbox}
\definecolor{tagbordercolor}{HTML}{B0BEC5}
\definecolor{tagbgcolor}{HTML}{ECEFF1}

\newtcbox{\captag}{nobeforeafter, colframe=tagbordercolor,
colback=tagbgcolor, boxrule=0.5pt, arc=1pt,
 boxsep=0pt,left=2pt,right=2pt,top=1.5pt,bottom=2pt,tcbox raise base}

\begin{document}
\title{\system: Topology-Driven Visual Analytics for Comparing Local Explanations}

\author{%
    Parikshit Solunke,
    Vitoria Guardieiro,
    João Rulff,
    Peter Xenopoulos,
    Gromit Yeuk-Yin Chan,
    Brian Barr,
    Luis Gustavo Nonato, and
    Claudio Silva
      \thanks{
  	Parikshit Solunke, Vitoria Guardieiro, João Rulff, Peter Xenopoulos, Gromit Yeuk-Yin Chan, and Claudio Silva are with New York University.
  	E-mail: {parikshit.s, vg2426, jlrulff, xenopoulos, gromit.chan, csilva}@nyu.edu.
    
        Brian Barr is with Capital One.
  	E-mail: brian.barr@capitalone.com.

        Luis Gustavo Nonato is with ICMC-USP, São Carlos, Brazil.
        E-mail: gnonato@icmc.usp.br. 
}
}




\maketitle
\begin{figure*}[ht]
  \centering
  \includegraphics[width= \linewidth]{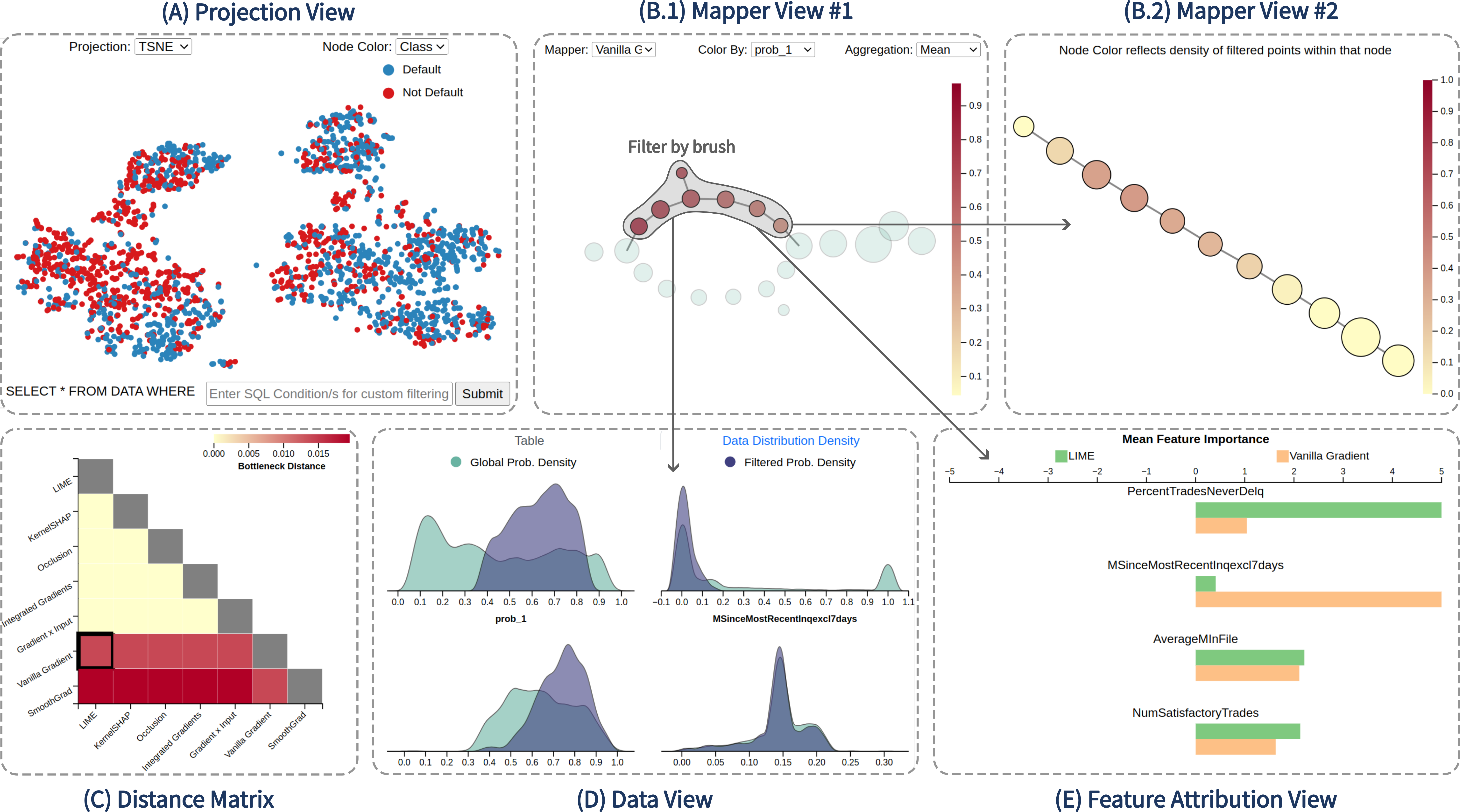}
  \caption{We use Mountaineer to compare black-box Machine Learning (ML) model explanations for the real-world HELOC dataset\cite{helocdata}.
  The Projection View (A) shows the original data projected into two dimensions. The user can choose among three different projection algorithms. The Distance Matrix (C) summarizes the distance between the topology of the explanation methods. When the user selects a cell in the matrix, the Mapper Views \#1 and \#2 (B.1, B.2) update to show the corresponding graph representations. In those views, the user can select nodes that they want to investigate (B.1). Then, the other graph coloring is updated to show the density of the selected samples (B.2). The Data View (D) presents the distribution of the features for the selected observations (in purple) and all observations (in green), arranged in descending order of difference between the two. The Feature Attribution View (E) displays the importance values for each feature according to the selected methods in decreasing order of importance. We can infer from the feature importance view that for the selected regions, there is a significant disagreement on feature importance between the two selected explanation methods.
  }
  \label{fig:teaser}

\end{figure*}

\begin{abstract}
With the increasing use of black-box Machine Learning (ML) techniques in critical applications, there is a growing demand for methods that can provide transparency and accountability for model predictions. As a result, a large number of local explainability methods for black-box models have been developed and popularized. However, machine learning explanations are still hard to evaluate and compare due to the high dimensionality, heterogeneous representations, varying scales, and stochastic nature of some of these methods. Topological Data Analysis (TDA) can be an effective method in this domain since it can be used to transform attributions into uniform graph representations, providing a common ground for comparison across different explanation methods. 
We present a novel topology-driven visual analytics tool, Mountaineer, that allows ML practitioners to interactively analyze and compare these representations by linking the topological graphs back to the original data distribution, model predictions, and feature attributions. Mountaineer facilitates rapid and iterative exploration of ML explanations, enabling experts to gain deeper insights into the explanation techniques, understand the underlying data distributions, and thus reach well-founded conclusions about model behavior. Furthermore, we demonstrate the utility of Mountaineer through two case studies using real-world data. In the first, we show how Mountaineer enabled us to compare black-box ML explanations and discern regions of and causes of disagreements between different explanations. In the second, we demonstrate how the tool can be used to compare and understand ML models themselves. Finally, we conducted interviews with three industry experts to help us evaluate our work.

\end{abstract}


\section{Introduction}

Black-box Machine Learning (ML) methods are being increasingly employed in critical and mainstream applications in industries such as healthcare, finance, transportation, and cybersecurity \cite{9237327}. While these complex yet opaque methods have led to significant performance boosts, there has been a rising demand for methodologies to explain model predictions. In particular, growing public concerns, as well as governmental regulations like the General Data Protection Regulation (GDPR), have propelled the urgency and desire for understanding ML model inner workings \cite{Voigt2017TheEG}.

As a result, a plethora of eXplainable Artificial Intelligence (XAI) \cite{confalonieri2021historical} techniques have been developed to \textit{locally} explain black-box ML models. These methods are local in the sense that they accept an observation as input and output a corresponding real-valued set. Typically, each value in the set serves as the feature attribution for the input observation. These feature attributions correspond to the importance level of a feature towards the model's prediction for that particular observation. Thus, at a high level, these techniques are intended to highlight the sensitivity of an observation to a given feature in the neighborhood around an observation. For instance, consider the case where a credit risk prediction model has assessed a loan applicant to be ``risky''. In such a scenario, an explanation technique would be expected to have greater attributions for features like credit score or annual income.

Working with local explanation methods is cumbersome for a variety of reasons. First, many local explanation methods often have their own hyperparameters that can significantly impact the resulting explanations. Furthermore, various local explainability methods have varying foundations and techniques for determining the model's behavior \cite{arrieta2020explainable, minh2022explainable} which can produce explanation outputs with varying interpretations and scales. Thus, it is common for different explanation methods to disagree \cite{krishna2022disagreement}. Lastly, there does not exist a universally optimal explanation technique. As a consequence, there is a pressing need for approaches to compare explanations of different XAI methodologies.

As ML models leverage increasingly large data sets, it is prohibitive for one to manually compare sets of explanations. In a recent technique called GALE \cite{pmlr-v196-xenopoulos22a}\cite{xenopoulos2022topological}, the authors propose using Topological Data Analysis (TDA) to summarize heterogeneous sets of local explanations for the binary classification problem. TDA \cite{wasserman2018topological} techniques use algebraic topological concepts to analyze the shape and structure of high-dimensional data. Given a set of local explanations, the authors use the Mapper algorithm \cite{singh2007topological} to build a graph that is a skeleton representation of the explanations' topology. In this graph, each node is a cluster of observations that have similar predicted probability and also similar explanations. The edges are built between nodes with observations in common. Thus, GALE provides a quantitative way to globally compare local explanations.

In addition to comparing explanation methods globally as proposed by GALE, we can utilize the graph representations to delve deeper into the variations between different explanation methods. For example, if one representation merges a set of observations into a single node, while another separates them into multiple ones, it would indicate that the explanation methods differ in their interpretations of those observations. Consequently, we can evaluate the feature distributions of those samples to determine a disagreement region between the methods --  that is, for which feature values the methods disagree. Also, we can inspect the actual explanations that the methods provide for those observations, determining how they disagree. Those conclusions are valuable from an explainability point of view since they allow us to discover regions of the feature space the explanation methods agree in, so we have more confidence in the validity of our explanations, and the regions that they disagree in, raising the possibility of one (or more) methods to be mistaken. Furthermore, the summarization of explanation attributions for regions of observations allows domain experts to ascertain which explanation method is more in tune with the reality of the problem's domain. However, such analysis requires manual comparison of the topological representations and the relationships between the graphs, observations, explanations, and model outputs.

In this work, we present \system, a visual analytics framework to analyze and compare ML explanation methods using techniques from Topological Data Analysis. We build upon the GALE methodology to provide a visual and interactive approach to compare local ML explanations. We design and develop \system in collaboration with industry ML practitioners who routinely work with black-box ML methods and XAI methods. \system is implemented within Jupyter Notebooks, which enables easy deployment and collaboration in a wide variety of ML environments. The primary contributions of our work include: 
\begin{itemize}
\item \textbf{\system}- a Jupyter Notebook-based visual analytics system that allows the comparison of ML model explanations by linking the topologies of explanations back to the original data and model predictions through multi-view and complex interactions and filtering capabilities.
\item A methodology for determining and analyzing regions of disagreement between local explanation methods based on their topological graph representations. 
\item
An evaluation of {\system} through case studies on real-world datasets that demonstrate the effectiveness of our TDA-based framework in visually guiding the comparison of ML model explanation methods.    
\end{itemize}

\section{Related Work} \label{sec:related-work}

\subsection{Visual Interpretability of Black-Box ML Models}
Human interpretations of ML model predictions commonly rely on visualizations of model properties and parameters. These interpretations can be broadly categorized into two groups - interpretations of white-box models and interpretations of black-box models. The structures in white-box models are self-explanatory and are usually discernible to human reasoning processes. For instance, the tree structure of decision trees provides a clear decision rule at every node and each branch represents the possible outcomes at that node. Therefore, decision trees can be easily understood through simple visualizations like flowcharts \cite{van2011baobabview}. As the rule list is essentially a set of IF-THEN statements, they can also be visualized through matrix visualizations \cite{ming2018rulematrix}. Additionally, linear models like generalized additive models can be visualized as a set of regression line charts showing feature importances at different ranges \cite{hohman2019gamut}.

Black-box models, on the other hand, are much more opaque to direct human interpretation \cite{von2021transparency} owing to a lack of access to model internals. Hence, approaches to interpreting black-box model predictions are often model agnostic. For instance, we can visualize the relationship between an attribute and prediction with a Partial Dependence Plot (PDP) by marginalizing the output with all other attributes in the dataset \cite{friedman2001greedy}. Given the rising need for transparency in ML predictions, there has been a significant amount of work in generating local explanations for model predictions \cite{arrieta2020explainable, minh2022explainable}. These explanation methods are usually attribution methods, which assign credits to each input point's features when they assess these features as being important for the ML model to predict the input’s outcome. They take the ML model and an instance as inputs. Then, they output a feature vector where the values indicate the importance of each feature to the ML model of that particular instance. 

LIME \cite{10.1145/2939672.2939778} is a popular method that computes attributions by adding perturbations to an input instance and training a linear classifier on these perturbed inputs to extract the locally important features. Another popular technique, SHAP \cite{NIPS2017_8a20a862}, uses shapely values from game theory to calculate the contribution of each feature of an input instance to the predicted value. For neural networks, the interpretations on the input can be retrieved by comparing the neuron activation differences between a baseline and the input so that inputs like images can be interpreted with a saliency map \cite{shrikumar2017learning, pmlr-v70-sundararajan17a}. 

These feature attributions can then be visualized with bar graphs. The feature attributions of different methods are on different scales, can vary significantly \cite{krishna2022disagreement}, and are often calculated stochastically. Additionally, attributions are computed with a local frame of reference - i.e. they're computed for every observation individually devoid of a global point of view. This raises two important questions - \emph{1) How do we evaluate the local explanations and compare different techniques at a global level}? and \emph{2) Given a set of explanations- how do we determine which are most suited for the problem on hand}?  

Our topology-driven framework focuses on improving the workflow of black-box model interpretations, in which 
practitioners need to evaluate different explanation methods to choose the best one for achieving their goals. Furthermore, our approach also focuses on allowing comparison between explanations at various granularities - ranging from a single observation to the global space.

\subsection{Topological Data Analysis}

Topology has been proven to be immensely valuable in visual data analytics applications, particularly in scientific visualization. Singh et al. \cite{singh2007topological} proposed the Mapper algorithm to transform multi-dimensional data spaces into simple topological skeletons enabling visualization and analysis. Large-scale 3D objects can be transformed into a simpler set of important nodes and edges that are faithful representations of the underlying data. These simplified representations allow systems to speed up rendering time \cite{lukasczyk2020localized}, segment 3D objects (e.g., different fish in a high-resolution CT scan) \cite{bock2017topoangler}, and identify dynamics in 3D simulations \cite{lukasczyk2017viscous}. TDA has also been used in Affective computing to visually compare human emotions \cite{elhamdadi2021affectivetda}. Pheno-Mapper \cite{zhou2021pheno} allows for the interactive exploration of Phenomics data guided by the topological summary of the manifold. 

In recent years, topology is being increasingly used for the goal of understanding and exploring Machine Learning Algorithms. In 2009, Carlsson \cite{carlsson2009topology} hypothesized that data is sampled from an underlying manifold - in other words, that data has ``shape''. Instead of looking at data and data distributions from a purely statistical angle, TDA unlocks a new paradigm of data analysis by studying the underlying manifold shape in an algebraic way. The introduction of software libraries such as GUDHI \cite{maria2014gudhi} and giotto-tda \cite{tauzin2021giotto} have significantly contributed to the growing adoption of TDA. 

There has been an uptick in the application of topology in the domain of interpretability of black-box ML models. TopoAct \cite{rathore2021topoact} presents a visual analytics system to study the topological structure of neuron activations in neural networks. Wheeler et al. \cite{Wheeler_2021} showed that topology can be used to study the activation landscapes i.e. how data transforms as it passes through the layers of a deep learning network.

Xenopoulos et al. \cite{pmlr-v196-xenopoulos22a}\cite{xenopoulos2022topological} proposed GALE, a framework to help identify topological differences between local explanation methods and identify appropriate parameters for local explainability methods. They employed the Kepler Mapper \cite{KeplerMapper_JOSS} library - which is a Python implementation of the Mapper algorithm to obtain the topological graphs of an input dataset. Additionally, they implemented the bottleneck distance metric \cite{cohen2005stability} to compare topological graphs of a set of explanation methods. However, GALE does not provide an interactive interface to study the topological graphs at different granularities and does not allow for on-the-fly exploration or linking of the topological graphs with the original dataset and feature attributions. In this work, we address those limitations by proposing a methodology to analyze the differences in the topology of different explanation methods and developing a tool to visually guide this exploration. In addition, we introduce a new strategy for automatically selecting the mapper parameters with theoretical guarantees, and that significantly decreases the time required for generating the graph representations.

While topology is gaining popularity in machine learning and visual analytics, there has been little work on the interactive visualization and analysis of topologies of ML explanations. With \system, we aim to bridge this gap and present a system that harnesses the qualities of topology in combination with visual analytics to tackle the problem of comparing local black-box ML explanation methods.
\section{Topological Background} \label{sec:tda}

\begin{figure}
    \centering
    \includegraphics[width=\linewidth]{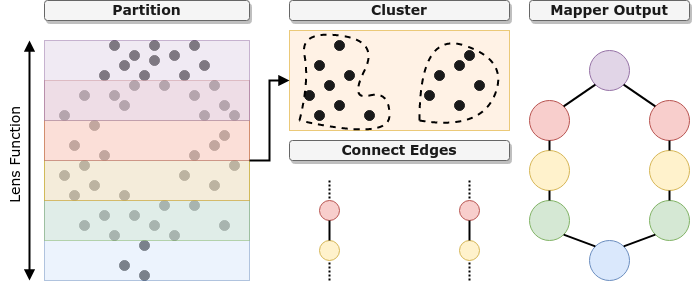}
    \caption{Mapper algorithm used to create an approximate Reeb graph. The input space is first divided into overlapping intervals based on lens function values. Then, the points within the intervals are clustered into nodes. Subsequently, edges are constructed between clusters that have common input points. Thus, the Mapper output is generated as a node-link graph forming a skeletal representation of the input space.
   }
    \label{fig:mapper}

\end{figure}

\subsection{Reeb Graphs and Mapper Algorithm}

Reeb graphs\cite{reeb1946points} are used in topology to analyze and understand the shape and structure of high-dimensional manifolds. Specifically, it is a graph that summarizes the topological structure of a continuous function on a given space. Consider a scalar function \(f:\mathbb{M}\rightarrow\mathbb{R}\). The level set \(f^-1 (a)\) at a given scalar value a is the set of all points that have the function value $a$. The Reeb Graph is computed by contracting each connected component of the level sets to a single point producing a skeleton-like representation. In essence, the goal is to identify the critical points of the function on the level set and connect them, with vertices representing the critical points and edges representing the connected components of the level set. Thus, it provides a compact representation of the topology of complex datasets and can be used to identify important features and relationships within the data.

However, in the real world, datasets are not continuous manifolds but are usually sets of high-dimensional data points\cite{carlsson2009topology}. Mapper\cite{singh2007topological} is a topological technique that can transform multidimensional input data into a representative graph using partial clustering. As shown in Figure \ref{fig:mapper}, the Mapper algorithm approximates a Reeb graph on a user-defined scalar function (also known as the \emph{lens} or \emph{filter} function) of a dataset. First, the algorithm partitions the data into overlapping subsets, or bins, based on the lens function values. Then, it constructs a graph that represents the relationships between the bins. The graph nodes represent the bins, and the edges correspond to the overlap between the bins. The resulting simplified graph can then be used to study the topology of the input dataset in a lower-dimensional space, increasing its interpretability and making it more conducive to being visualized.

\subsection{Mapper Parameters}

In addition to the input space and the scalar lens function, the Mapper algorithm requires three parameters - 1) the \emph{resolution} specifies the number of intervals the range of the scalar function is divided into, 2) the \emph{gain} defines the amount of overlap between consecutive ranges, and 3) the clustering algorithm (which may need its own hyperparameters) used to perform clustering. These parameters are crucial for the structure of the Mapper graph. A high resolution or a low gain may result in a loosely connected graph with a very high number of nodes. On the contrary, a high gain may result in too many interconnected nodes while a low resolution might result in very few nodes.

A Mapper graph constructed from ill-suited parameters generates an unfaithful and uninsightful representation of the underlying input space. Thus, to enable accurate analysis, it is essential to ensure that the parameters are properly selected. In this work, we propose an automatic parameter selection approach based on the \emph{stability} of the resulting graph, which is presented in Section \ref{sec:parameter-selection}.

\subsection{Topological Persistence and Bottleneck Distance}

The topological persistence of a space measures the robustness of its topological features as it transforms. To obtain the persistence, we begin by constructing the filtration of the space, which can be thought of as a sequence of nested sub-spaces that describes the topological features of the space at different levels of detail. In the case of the Mapper representations, the filtration function is the lens or filter function $f$ provided for generating the Mapper. The filtration is the set of sublevel sets $f^{-1}(-\infty, a)$, for all $a$ on $f$'s domain. As the function value $a$ increases, the topology of the sublevel sets changes at the critical points of the function \cite{pmlr-v196-xenopoulos22a}\cite{xenopoulos2022topological}. Essentially, at critical points, a new topology (a $k$-dimensional cycle, such as a connected component or a loop) is either created or destroyed. For a topological feature, the topological persistence can be defined as \(f(c_j)-f(c_i)\) \cite{edelsbrunner2000topological} where \(c_j\) denotes the critical point where the topological feature was destroyed and \(c_i\) denotes the critical point where the feature was created. Therefore, the topological persistence of a feature can be understood as an indicator of its lifetime.

The birth and death times of these topological features can then be plotted as 2D scatter plots called persistence diagrams\cite{edelsbrunner2000topological}. This diagram can then be used for comparing topologies. 
The bottleneck distance\cite{cohen2005stability} is frequently used to measure the difference between two persistence diagrams\cite{otter2017roadmap,chazal2021introduction}. This distance is measured as follows: (1) every point from a persistence diagram is paired with another point from the other diagram, (2) if a point is not paired, it is paired to the diagonal of the diagram, (3) the bottleneck distance is the maximum sup norm between the pairs. A smaller bottleneck distance indicates greater similarity between the two topologies. This distance is frequently used because it is stable to small topological perturbations \cite{cohen2005stability}. In our work, we use the bottleneck distance as the distance measure to indicate the similarity between topological graphs of ML explanations. 

\section{Design Requirements} \label{sec:design}
\system was designed through an iterative process with regular consultation and input from industry practitioners and domain experts. Owing to an ever-increasing number of local explainability methods at their disposal, practitioners are often faced with the challenge of assessing and selecting the most appropriate method for the problem at hand. This task requires a global perspective on methods that are by design, local. Our practitioners found the proposition of using TDA to provide this global perspective to be promising. However, they identified certain challenges that would need to be addressed for TDA to be of utility in this domain. 

Firstly, they pointed out that the graph by itself only represents the underlying structure of the explanation space but conveys no additional information about the model, the observations, or the explanations. The topological representation also does not encode information about the actual feature values of the dataset. Furthermore, the parameters of the Mapper algorithm can greatly affect the structure of the resulting graph, and thus it is necessary to ensure that the differences between graphs arise from differences in explanations and not due to poor parameter choice. Finally, our practitioners pointed out the need to be able to compare feature attributions for smaller regions of interest within the dataset and indicated that they conduct their work using popular packages in Python within Jupyter Notebooks.

Based on discussions with domain experts and a survey of existing literature, we formulated the following design requirements for our system:

\begin{itemize}[label=C\arabic*.,itemsep=0mm, topsep=2mm]

\item[\textbf{R1.}]
\captag{\sffamily \textsc{\small{Visualization}}}
\textbf{Encoding relevant information in the Mapper graph's nodes and edges.}
The output of the Mapper algorithm is a simple graph, with a list of nodes and edges. However, its base output does not take advantage of visualizing through the various visual channels available. We could boost the utility of the Mapper graph by encoding information about the corresponding data points in the node sizes and colors. The user should also have control over the information being displayed in the visual encodings as well as the aggregation methods being used.

\item[\textbf{R2.}]
\captag{\sffamily \textsc{\small{Interaction}}}
\textbf{Linking the outputs of topological data analysis with familiar visualizations and interactions.}
A topological representation is not intuitive on its own, and in a sense is a clustering of the input space. However, by linking it to the projection and data views with interactions such as brushing and querying, we can make the interface highly intuitive and insightful. It is also essential to empower the end-user to study and summarize regions of the graphs. 


\item[\textbf{R3.}]
\captag{\sffamily \textsc{\small{Comparison}}}
\textbf{Comparing different Mapper outputs visually.}
GALE provides a distance metric to compare graphs obtained from the Mapper algorithm. However, it does not support visual and interactive comparisons between those graphs. The tool must allow the user to compare two graphs in terms of their global structure while also allowing comparison between smaller regions of the graphs.

\item[\textbf{R4.}]
\captag{\sffamily \textsc{\small{Comparison}}}
\textbf{Aiding comparison of feature attributions across explanation methods.}
Different explanation methods generate explanations in different dimensions. Thus, two methods can output explanations that are on distinct scales but actually convey the same information about the importance of each feature. The tool must allow the user to compare feature importance across various explanation methods globally as well as for smaller regions of the dataset.

\item[\textbf{R5.}]
\captag{\sffamily \textsc{\small{Integration}}}
\textbf{Providing functionality in a notebook environment to support ease of use.}
Most ML practitioners are familiar with and conduct their work using Jupyter Notebooks. To encourage collaboration and wider adoption of \system, it is vital to bring the visual analytics tool to where the data is, i.e. the tool must be implemented within this environment to maximize its utility.

\end{itemize}

\section{\system} \label{sec:mountaineer}

\begin{figure}[t]
    \centering
    \includegraphics[width=0.9\linewidth]{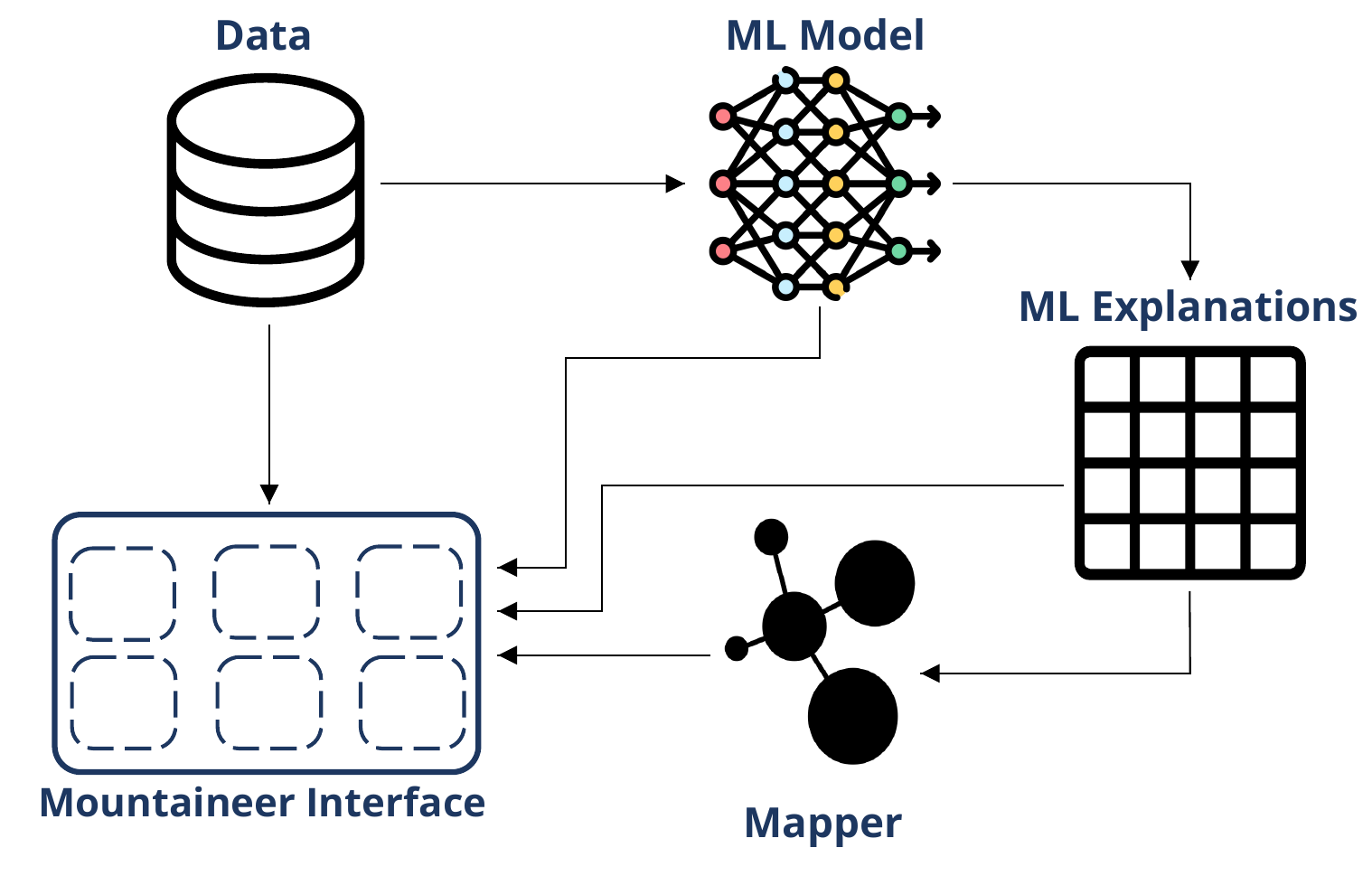}
    \caption{\system links the data, model predictions, chosen explanation results, and their topological graphs into an interactive visual framework.}
    \label{fig:Architecture}

\end{figure}

In this section, we describe {\system}, a visual analytics system that allows topology-driven interactive comparison of ML explanation methods. We begin by detailing the process we used in our framework to create topological graphs and the methodology for selecting the Mapper parameters. Next, we explain how we refine the topological graphs. Then, we briefly describe the multiple views in {\system}, shown in Figure 1, and outline the interactions that the system facilitates. We elaborate on how the multi-views and interactions address the design requirements identified in Section~\ref{sec:design} and finally, we discuss the implementation of {\system}.

\subsection{Creating Topological Representations}

Recall that ML model explanation methods output a feature attribution vector \(\mathbb{R}^k\) for each input point \(\mathbb{R}^d\). The feature attributions reflect the “importance” of each feature based on the assessed impact of that feature on the model's prediction for the given point. Thus, a given explanation approach E can be thought of as a high-dimensional function \(f_E:\mathbb{R}^d \rightarrow\mathbb{R}^k\) where k = d for most approaches. 

We create topological graphs of the explanation attributions using the Mapper algorithm. For each of the explanation methods that we wish to compare, we construct a graph using the feature attributions of each observation as the input space. As the scalar lens function, we use the predicted probability for the target class of our binary classification model. We use agglomerative clustering as our clustering method. Hence, in addition to the resolution and gain parameters, we also need to provide a distance threshold hyperparameter. Our motivation for employing this clustering method is that it enables us to automatically select the appropriate parameter, as discussed in Section \ref{sec:parameter-selection}, where we present our parameter selection strategy. Nevertheless, it is possible to use alternative clustering methods, such as DBSCAN, and employ different parameter selection strategies by providing \system with any MapperComplex object generated using the GUDHI library~\cite{gudhi:urm}.

\subsection{Parameter Selection} \label{sec:parameter-selection}

In this work, we rely on the findings presented by Carriere et al.\cite{carriere2018statistical} to guide our selection of gain and clustering parameters. They determined that the ideal gain value should be selected within the range of $1/3$ and $1/2$, so we set it to $0.4$. The distance threshold parameter of the agglomerative clustering is estimated by taking multiple subsamples of the point cloud of explanations. Then, we calculate the distance between the point cloud and each subsample using the Hausdorff distance. The average Hausdorff distance between these subsamples is set as the distance threshold. For further information on the parameter selection procedure, we refer the reader to Carriere et al.\cite{carriere2018statistical} and GUDHI reference manual \cite{gudhi:urm}.

Finally, to select the resolution parameter, their analysis relies on a regularity property of the filter function that is not guaranteed to be true in our case. So, we select the resolution that generates the most stable mapper, as in Xenopoulos et al.\cite{pmlr-v196-xenopoulos22a}\cite{xenopoulos2022topological}. The motivation for using stability lies in the notion that under ideal circumstances where parameters are accurately estimated, the mapper should exhibit minimal changes when generated with resamples of the point cloud. So, we employ bootstrapping to calculate the bottleneck distance between two mappers: one built with the selected parameters and the original point cloud, and the other with the same parameters but using resamples of the point cloud. We then perform a grid search across a range of possible resolution values and select the one that maximizes stability while using the gain and clustering method determined earlier.

\subsection{Reducing Clutter in Mapper Output}

The Mapper algorithm identifies connected components with the use of overlapping ranges. This characteristic of the algorithm can lead to two different nodes of the graph having the same set of input points. Consequently, there can be redundant nodes and edges in the graph which cause the displayed graph to have unnecessary clutter. Thus, it is essential to prune the output before displaying it. We employed graph summarization \cite{navlakha2008graph}, which reduces components in the graph by combining similar nodes or edges. In our case, given a set of nodes V where each node corresponds to a set of inputs of our dataset, we aim to find a partition P such that the number of nodes is minimized by combining nodes that have a similar set of inputs above a certain threshold into one single node. Thus, we can formulate the goal as minimizing an objective function as follows:
\begin{center}
minimize \(\parallel P \parallel\) \newline
subject to \(\forall v_i,v_j \in V \rightarrow p_i \in P \)  where sim(\(v_i,v_j\)) \(\geq\) 0.9\
\begin{figure}[t]
    \centering
    \includegraphics[width=\linewidth]{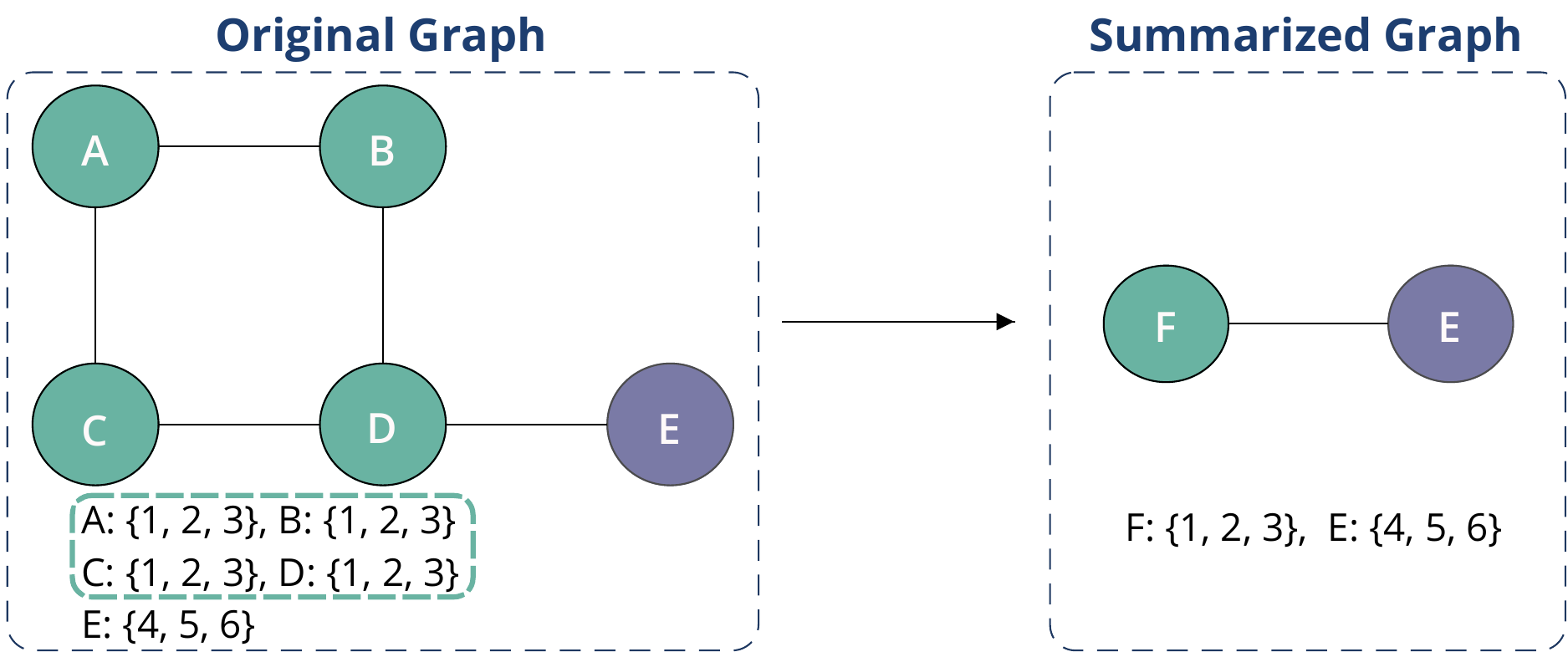}
    \caption{Graph summarization eliminates visual clutter by reducing redundancy. Nodes A, B, C, D (original graph) form a connected component and have the same data. Thus, they are aggregated into a new node F (optimized graph).}
    \label{fig:summarization}

\end{figure}
 
\end{center}

Here sim represents the Jaccard similarity between two node clusters. To compute the optimized graph, we construct a hierarchy of node clusters with Jaccard distance and single-linkage criterion and then form the optimized clusters based on the distance threshold in the hierarchy. In the case of topological graphs, such summarization removes not only the redundant nodes but also the excessive edges formed between very similar nodes. As shown in Figure \ref{fig:summarization}, similar nodes often form cliques among each other. Therefore, removing nodes with our objective function also implies removing clique edges in the graph. The high similarity threshold ensures that only redundant nodes are consolidated, resulting in a concise graph without removing potentially important nodes. Additionally, the tool offers the flexibility to disable summarization during the construction of mapper graphs, if required.

\subsection{Views}
\system consists of five linked views built to support interactive analysis of ML explanations at various levels of granularity. In this section, we briefly outline the views in our system.

\subsubsection{Projection And Query View (Figure 1A)}
The projection view shows a two-dimensional projection of the input dataset. \system supports t-SNE \cite{van2008visualizing}, UMAP \cite{mcinnes2018umap}, and PCA \cite{wold1987principal} projections - which are dimensionality reduction techniques popular for visualizing high dimensional data. The user can change the displayed projection on the fly as per their preference. We also allow the user to specify whether the input points are colored by their true class or by the correctness of the model's prediction. This perspective may help uncover disparities in class representation or identify discernible clusters, should they exist within the dataset. Notably, it would help highlight and investigate predominant or underrepresented groups, as well as clearly defined clusters of data points.

The SQL querying interface grants the user fine-grained control over the filtering and enables the examination of smaller regions of interest within the dataset. For example say a medical practitioner is trying to evaluate a diabetes prediction model and wants to look only at how the explanations are performing for ``high-risk'' patients - those who are above the age of 50 and have high glucose concentrations, the practitioner can easily add the conditions to the query to observe the region of interest and can also modify these queries to obtain results on the fly. 

\subsubsection{Mapper View (Figure 1 B.1 and B.2)}
The Mapper view consists of side-by-side representations of two node-link graphs corresponding to the explanation methods selected by the user. We can think of the Mapper output as an overlapping clustering of the explanation space, the nodes represent the clusters in the explanation space and there exists an edge between any two nodes if they share common input points. This helps to identify the topological structure of the explanation space and the relationships between the clusters as well as compare these structures and relationships for two different graphs (\textbf{R3}).    

The Mapper algorithm outputs a graph representing the topological skeleton of the input space. However, as the explanation space has a one-to-one relationship to observations in the input dataset, we can link these nodes back to the original dataset and encode a significant amount of information in the nodes making our graphs much more insightful (\textbf{R1}). Firstly, we use the radius of the nodes as a visual encoding representing the number of points within a cluster. Larger the radius of a node, the more points belong to that cluster. Secondly, we enable the user to select the attribute the nodes are being colored by - which includes all of the features in the data space as well as the prediction probabilities and true class values. Additionally, the user can also select the type of aggregation being used to color the nodes. The aggregations we support include mean, median, max, min, and standard deviation of the selected attribute. Encoding information about the input dataset in the topological graph of the explanation space makes \system a powerful tool to study and analyze ML model explanations.

\subsubsection{Distance Matrix (Figure 1C)}
The distance matrix displays a heat map of the distance between the topological representation of the explanations. We use the bottleneck distance\cite{cohen2005stability, pmlr-v196-xenopoulos22a, xenopoulos2022topological} as the measure to calculate the distance between topological graphs. The bottleneck distance is the ``cost'' of the optimal matching between points of two persistence diagrams i.e. the minimum ``cost'' required to transform one graph into another, where the cost is determined by the pairwise distance between points. In essence, the explanation topologies which most resemble each other will have the lowest bottleneck distances. 

The user can click on any cell in the heat map to select the explanation methods to compare. On selection of new methods, the Mapper View updates with graphs corresponding to the chosen explanation methods. The Feature Attribution view will also update to show the feature attributions for the selected pair of explanations. Thus, the heat map provides an overview of all of the explanation topologies and provides a starting point for analysis. 

\subsubsection{Data Table and Distribution View (Figure 1D)}
The data view consists of two tabs that the user can switch between - one for the raw data table of the dataset and another showcasing the distribution of feature values.

\paragraph{Data Table} The table lists the feature values for all observations in addition to their predicted probabilities and their true class values. Additionally, we also show the average value of the entire dataset. Whenever the user queries for a region of interest, we also display the feature averages for the filtered points and highlight the difference between the local and the global averages (\textbf{R2}).

\paragraph{Distribution View} In the distribution view we show small multiples of density graphs of Kernel Density Estimations (KDE) for all of the features of the dataset as well as the predicted probabilities and the actual class values. Whenever the user queries or brushes a region of interest in the other views, we compute the KDE graphs for every feature for the selected region and overlay these on top of the corresponding graph of global distribution for each feature, as shown in Figure \ref{fig:teaser} (\textbf{R2}). Furthermore, we order these graphs by placing the features with the largest differences between the distributions first, therefore highlighting the most prominent feature differences.

\subsubsection{Feature Attribution View (Figure 1E)}
The feature attribution view empowers the user to compare the explanation attributions for the two selected explanation methods. As feature attributions have varying scales, comparing the raw numbers for every feature between the explanations is not indicative of a difference in feature importance. Hence, we derived a relative feature importance level metric that scales the feature attributions for every explanation to a range between $-5$ to $5$ (\textbf{R4}). 

First, we take the maximum absolute attribution value for one particular explanation method and assign that value an importance level of $5$ for that explanation. Then, we scale the attributions for the same explanation to an importance level between $-5$ to $5$ with a negative value indicating a negative impact on the predicted probability for the target class. We repeat the above process for each explanation with its own maximum absolute attribution value being assigned an importance level of 5 thus ensuring that every explanation's attributions get scaled to importance ranges between $-5$ and $+5$.

We use bidirectional bar charts as the visual encoding for the mean feature importance levels. Additionally, to minimize the need to scroll, we display the feature attributions in descending order of absolute aggregate importance, thus ensuring that the most important features are always visible first.

\subsection{Linked Views and Interactions to Support Analysis}
We use this section to briefly outline the interactive workflows available to an end-user in \system to aid their analysis.

\subsubsection{Mapper View Interactions}
The user can brush any of the two Mapper graphs to highlight one or many nodes of interest. Whenever a graph is brushed, the data projection is updated to show the data points that fall within the selected nodes. Meanwhile, the color of the nodes in the other Mapper view is updated to reflect the densities of the data points within each node. The data table is updated to show the selected points meanwhile the distribution view is updated to show the density distributions of attributes for the entire dataset as well as the attribute distributions for the selected data points. The distributions are re-ordered in descending order of distribution difference. The Feature Attribution view is updated to show the mean importance level for all features. The attributions are re-ordered by descending value of combined absolute feature importance for both methods to highlight the most important features first (\textbf{R2}).

\subsubsection{Projection and Query View Interactions}
The user has two options to select a subset of points within the Projection and Query View. The projection view can be brushed directly to select a subset of points. Alternatively, SQL conditions can be entered in the query input to obtain fine-grained control over the data points being highlighted. The query view supports querying by any column in the dataset as well as the predicted probabilities and real class values of the data points. This enables the user to drill down to the exact set of points of interest than simply brushing the view.

Whenever a set of points are selected either by brushing or querying, the colors of nodes in both mapper graphs are updated to reflect the densities of the selected points within each node. Additionally, the Data Table, the Distribution Views, and the Feature Attribution respond to brushing in the same way as for brushing of the graphs and are re-ordered and updated to display the information about the selected points (\textbf{R2}).

\subsection{Implementation}

{\system} is a modern web-based system implemented as a Python library that fully integrates visual analytics in a Jupyter Notebook environment (\textbf{R5}). We intend to make \system available as a pip installable package. The front end is implemented using JavaScript, React, and D3.js\cite{6064996}. We create our graphs using the \emph{create\_mapper} method of the GALE library, which implements the Mapper algorithm\cite{singh2007topological}\cite{KeplerMapper_JOSS} to create topological graphs. This method takes the input space, lens function, resolution, gain, and clustering algorithm as arguments and returns an object detailing the node clusters and the links between these clusters. The back-end, which processes these Mapper outputs, generates the projections for the projection view, and eliminates graph redundancies is implemented using Python. To connect the front-end to the back-end and render the web-based system within a Jupyter Notebook, we use the library NotebookJS\cite{ono2021interactive}.

To use the tool, the user calls the \emph{visualize()} method of the Mountaineer class. The method requires the user to pass the dataset, the actual class values for every observation, the model predictions, and at least two explanation attributions and the corresponding Mapper outputs for each of these explanations as arguments to the \emph{visualize()} method. Additionally, the user can provide optional but recommended arguments like column names, explanation labels, and class labels to add to the readability of the results.

\section{Evaluation} \label{sec:use-cases}

\subsection{Case Study 1: Comparing Local Explanation Methods}

In this case study, we demonstrate how \system can be used to compare and contrast between ML model explanations on a real-world dataset. We use the Home Equity Line Of Credit (HELOC) dataset~\cite{helocdata}, which contains $9,871$ credit applications with $24$ continuous features. The task here is to predict whether an applicant would repay the credit. This dataset was also used by Han et al.\cite{NEURIPS2022_22b11181}, where the authors split the data into a training set with $80\%$ of the samples and a test set with the remaining $20\%$ of samples. Then, they trained a feed-forward neural network with three hidden layers with eight hidden nodes each. Here, we use the same neural network they trained as our prediction model with the following local explanation methods: Vanilla Gradient, Gradient x Input, Occlusion, LIME, KernelSHAP, SmoothGrad, and Integrated Gradients from the Captum Python library~\cite{kokhlikyan2020captum}. The parameters of the explanation methods were the same as Han et al.\cite{NEURIPS2022_22b11181}.

Using \system to compare those explanations, we first notice in the Distance Matrix view (Figure~\ref{fig:teaser}-C) that the first five explanation methods shown (LIME, KernelSHAP, Occlusion, Integrated Gradients, and Gradient x Input) have low bottleneck distance with each other. Comparing the corresponding graphs in the Mapper view (R3), all have similar line-like structures without branches or holes, such as the one shown in Figure~\ref{fig:teaser}-B.2. With the Feature Importance view, we observe that those methods have the same top 4 most important features, with the same importance sign and similar importance values (R4). Those features were the percent of previous trades that were not delinquent (PercentTradesNeverDelq), ExternalRiskEstimate, AverageMInFile, and NumSatisfactoryTrades. Meanwhile, the last two explanation methods (Vanilla Gradient and SmoothGrad) have higher bottleneck distances with all other methods, and their corresponding graphs have topological holes (Figure~\ref{fig:teaser}-B.1). They also disagree with the previous methods over the importance of many features while agreeing with each other. Three of their four most important features are distinct from the earlier methods' top 4: the trades with high credit utilization ratio (NumBank2NatlTradesWHighUtilization), months since the most recent credit inquiry (MSinceMostRecentInqexcl7days), and the number of credit inquires in the last six months (NumInqLast6M). 

\begin{figure}[t]
    \centering
    \includegraphics[width=\linewidth]{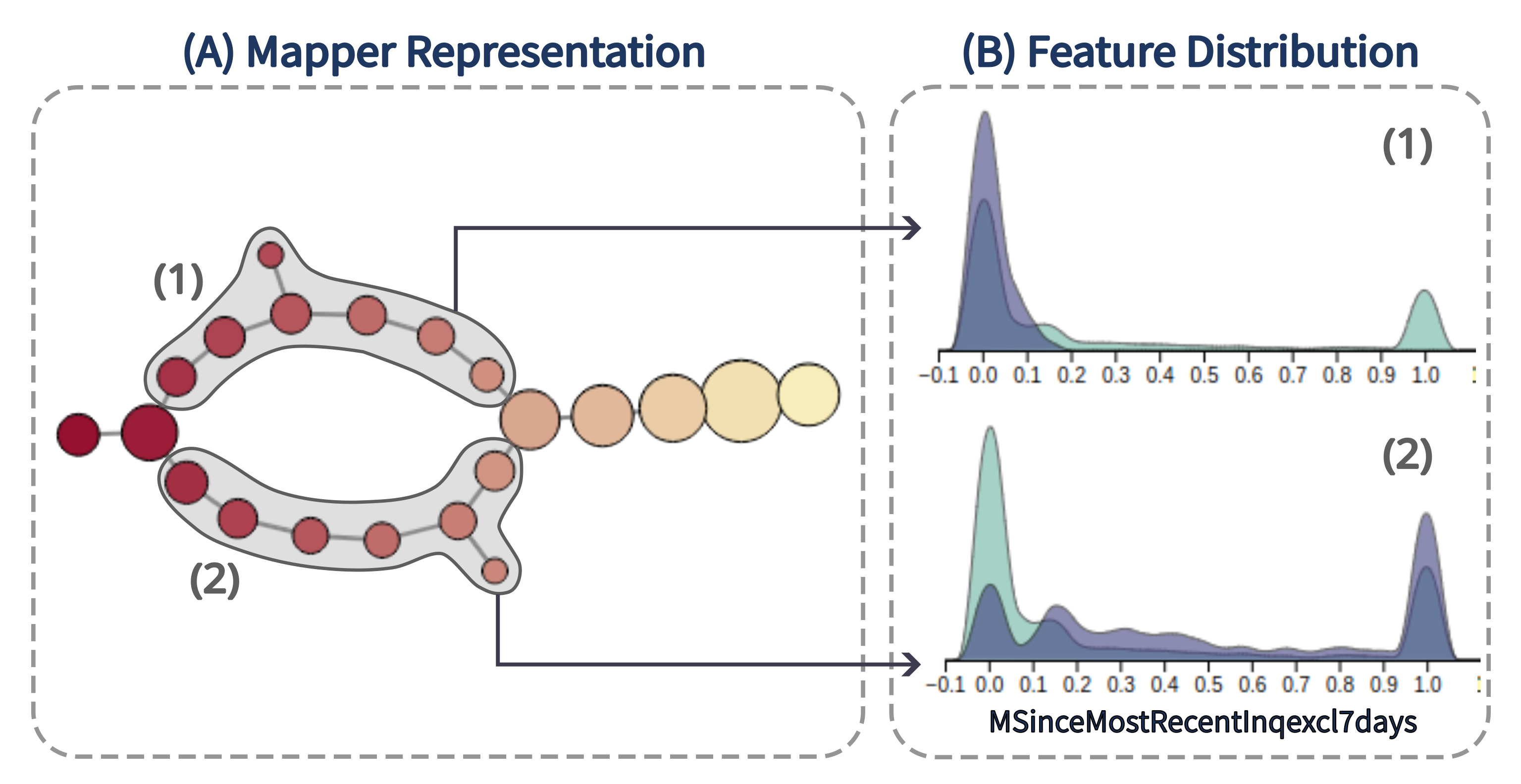}
    \caption{Mapper View (column A) and distribution of the feature \textit{MSinceMostRecentInqexcl7days} (column B) for Vanilla Gradient explanation method in Case Study 1. We select each ``side'' of the hole and analyze the feature distribution, concluding that one side corresponds to samples with low values for the feature and the other to high values.}
    \label{fig:casestudy1-1}

\end{figure}

Next, we investigate why Vanilla Gradient and SmoothGrad have topological holes. We start with Vanilla Gradient, which only has one hole. Using the interaction of the Mapper View, we select one side of this hole, as shown in Figure~\ref{fig:casestudy1-1}-A.1. We notice on the Data Distribution View (Figure~\ref{fig:casestudy1-1}-B.1) that the corresponding samples have low values for the feature MSinceMostRecentInqexcl7days, meaning that the client had a recent credit inquiry (R2). This is indeed the most important feature for those samples, having a positive importance. However, when we select the other side of this topological hole (Figure~\ref{fig:casestudy1-1}-A.2), we see that the samples have higher values and low importance for this feature (Figure~\ref{fig:casestudy1-1}-B.2). All other features have similar distributions and importance on both sides. Thus, according to Vanilla Gradient, the most recent credit inquiry made only matters when it was pretty recently. In this case, it is very relevant information for predicting whether the loan will be paid back. 

\begin{figure}[t]
    \centering
    \includegraphics[width=\linewidth]{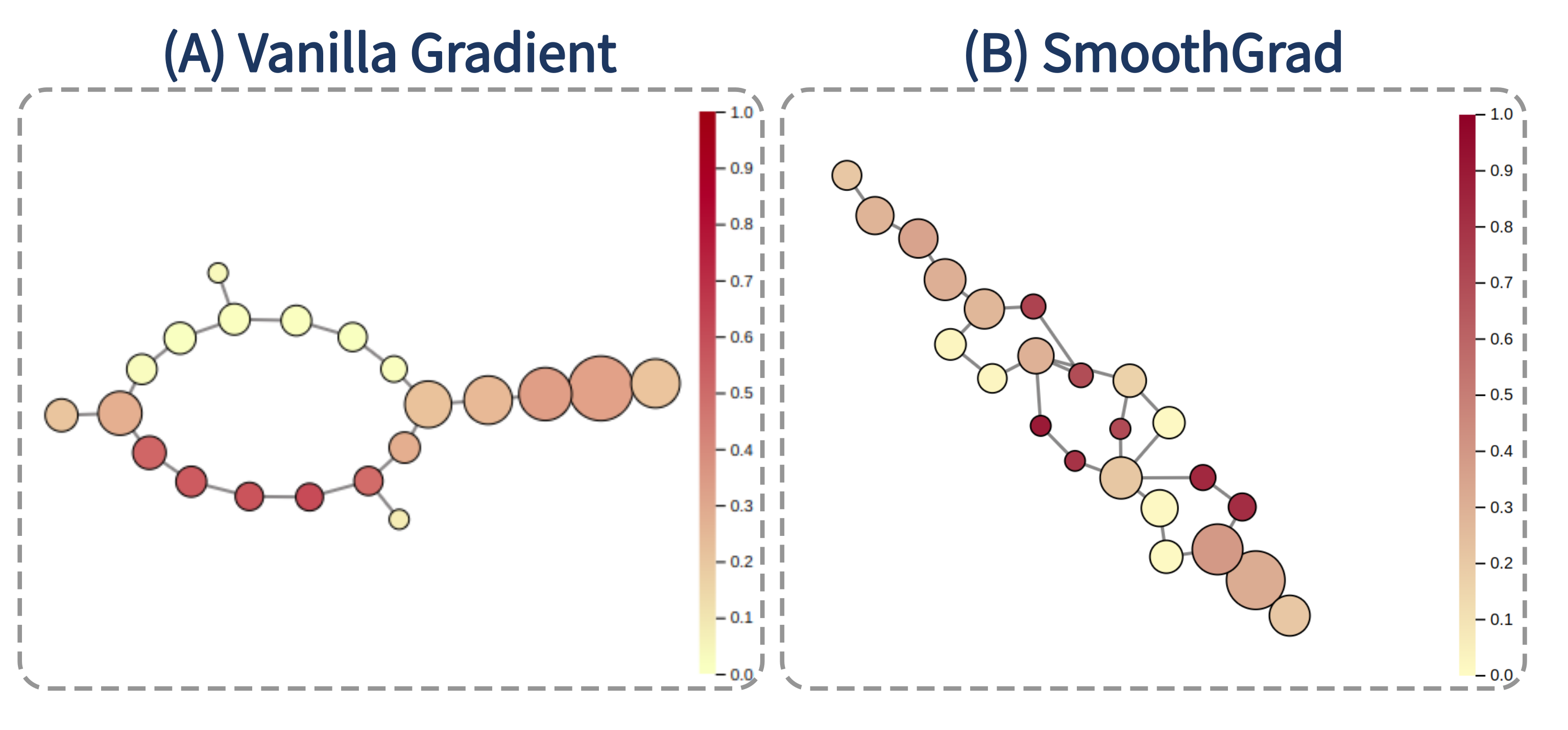}
    \caption{Vanilla Gradient (left) and SmoothGrad's (right) mappers with the nodes colored by the average value of feature \textit{MSinceMostRecentInqexcl7days}. All of SmoothGrad's topological holes have one side with low values for this feature and another with higher values. This is the same behavior found on Vanilla Gradient.}
    \label{fig:casestudy1-2}

\end{figure}

Following, we analyze whether the topological holes from SmoothGrad (Figure~\ref{fig:casestudy1-2}-B) are also due to the same reason. Unlike Vanilla Gradient (Figure~\ref{fig:casestudy1-2}-A), this method has three topological holes, but the predicted probability range for the holes is similar to the range of Vanilla Grad's hole. So, we use the Mapper View's interaction to change the node coloring to be the average value for the MSinceMostRecentInqexcl7days feature (R1). Figure~\ref{fig:casestudy1-2} shows the resulting mapper views. All holes have one side with a high average and a side with a low average for this feature. Investigating the two categories of sides, we notice the same behavior as for the previous method -- the last credit inquiry matters a lot when it was made recently and does not matter otherwise. However, it is relevant to understand why SmoothGrad has more holes than Vanilla Gradient. So, we analyze the nodes from SmoothGrad that correspond to nodes in different sides of Vanilla Gradient's hole and conclude that although MSinceMostRecentInqexcl7days is the most important feature for the samples with low values for it, SmoothGrad gives it a smaller importance than Vanilla Gradient. Also, SmoothGrad assigns higher importance to the other features for the samples with low and high values of MSinceMostRecentInqexcl7days. Therefore, the disagreement we notice between those two methods is only regarding the overall importance of this feature compared to the other features.

The results we found are consistent with the conclusions of Han et al.\cite{NEURIPS2022_22b11181}. In their work, the authors show that, for continuous data, the explanations generated with the methods LIME, KernelSHAP, Occlusion, Integrated Gradients, and Gradient x Input do not approximate the model's gradient but the gradient multiplied by the input. Meanwhile, Vanilla Gradient and SmoothGrad are indeed capable of approximating the gradients. Therefore, for our analysis to be coherent with their theoretical conclusions, we expect the first five methods to generate similar explanations distinct from those generated by Vanilla Gradient and SmoothGrad, which is exactly what we find with \system. In addition, \system is capable of detecting why the two groups disagreed -- a very important feature (MSinceMostRecentInqexcl7days) is only important when it has low values, so the methods that approximate the gradient times the input will not be able to detect this.

\subsection{Case Study 2: Comparing Models}

\begin{figure}[t]
    \centering
    \includegraphics[width=0.7\linewidth]{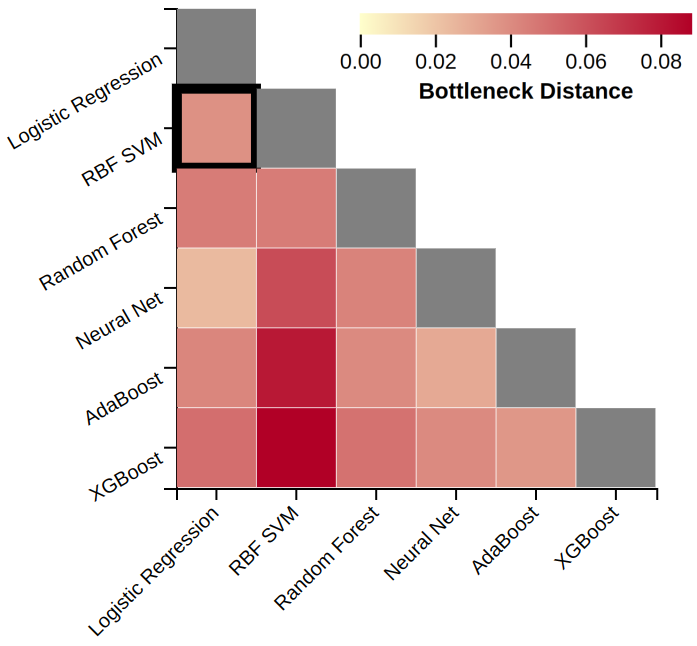}
    \caption{Distance Matrix for Case Study 2, we can see that the topologies of all LIME explanations differ from each other, with the closest being Neural Net and Logistic Regression.}
    \label{fig:casestudy2-1}

\end{figure}

\begin{figure*}[t]
    \centering
    \includegraphics[width=\linewidth]{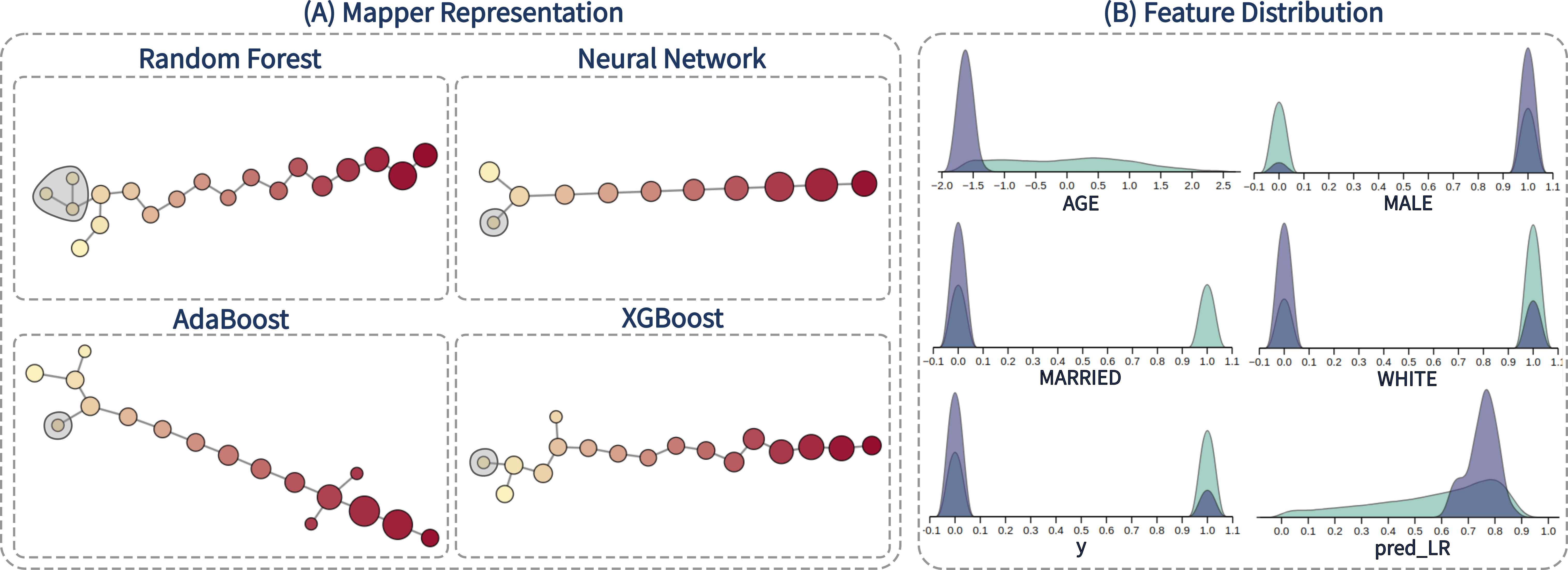}
    \caption{On the left (A) are the mapper graphs of Random Forest, Neural Network, AdaBoost, and XGBoost for Case Study 2. All mappers have a branch containing the same samples, which is highlighted on the graph. On the right (B) are the distribution of the most distinct features for those samples, corresponding to mostly white, young, and unmarried males. Most of those samples are unemployed (as shown on the y distribution). The selected models predict low probabilities for those samples (indicated by the node color), but Logistic Regression predicts high probability (shown on the pred\_LR distribution).}
    \label{fig:casestudy2-2}
\end{figure*}

An important motivation for comparing different explanations is to evaluate different models trained for the same task and with the same data. So, in this case study, we employ \system for this kind of comparison. We used the folktables~\cite{ding2021retiring} ACSEmployment dataset and selected a subset of the 2018 American Community Survey (ACS) containing information on 159,299 adults residing in the state of New York. The task is to determine whether a person is employed using features such as age, years of education, gender, relationship status, and others. There are both numeric and categorical features, with the majority being categorical. The dataset is balanced, with 57\% of the people being employed. We randomly split the data into 70\% for training and 30\% for testing. Then, we train multiple models using the same training set. The models trained and their corresponding accuracies are Logistic Regression (71.9\%), RBF SVM (76.4\%), Random Forest (76.2\%), Neural Network (76.7\%), AdaBoost (76.4\%), and XGBoost (77.0\%). For the ensemble-based (Random Forest, AdaBoost, XGBoost) classifiers, the predicted probability of classification is computed as the weighted mean predicted probability of the classifiers. Then, the probabilities are calibrated with an isotonic regression. For RBF SVM, the probabilities are estimated using Platt scaling. We use LIME to generate local explanations for all models with the exception of Logistic Regression, whose explanations were set as the model's coefficients. The number of perturbations is set to 1,000, ensuring that the explanations converge.

At first glance, the Bottleneck Distance view (Figure~\ref{fig:casestudy2-1}) shows that all models have different topologies, with the most similar being Logistic Regression and Neural Network and the most dissimilar pairs being RBF SVM with XGBoost, AdaBoost, and Neural Net. Inspecting the Graph view (Figure~\ref{fig:casestudy2-2}-A), we notice that all representations have a single component with a line-like structure (called a trunk), and most (except LR and SVM) have branches near either of the extremities (R3). By design, the LR has the same explanations for all samples (the model’s coefficients), so it will not have any branches or outliers. The main difference between SVM’s mapper and the others is that many of the samples are grouped into only two nodes – a consequence of the model predicting a quite high probability for many samples (R1, R2). The other models (especially XG, AB, and NN) have more evenly distributed probability distributions.

Another clear difference between the models can be noticed in the Feature Importance view (R4): for RF, AB, and XG, only the features of age and school years are on average important; for SVM and NN, age and school are also important, but so are other features such as cognitive difficulty (both models), if the person is naturalized in the US (SVM), and gender (NN); and for LR, the most important feature is cognitive difficulty, followed by age, hearing difficulty, and whether the person has been in the military – while school years is only the 7th most important feature.

Figure \ref{fig:casestudy2-2} shows that RF, NN, AB, and XG all have a low probability branch (Fig \ref{fig:casestudy2-2}-A) composed of samples of young, unmarried, under-average school years and mostly male persons (Fig \ref{fig:casestudy2-2}-B) (R2). Although all models assign negative importance to the age feature in the overall population, for this sample, the four models agree that age has positive importance and is the most relevant feature, with school years taking a distant second place in importance (R4). Almost all samples on those branches are unemployed. The LR model could not learn this pattern, classifying all those samples as employed. This branch is the only one for RF and NN, but AB and XG have more branches.

XG has another low-probability branch (R1), but its samples are all non-US citizens, female, married, young, and with few school years (R2). For those samples, age was the most important (and positive) feature, gender being the second most important (also positive) (R4). Even though no other model has a branch exclusively with a similar population cut, SVM, RF, and NN all agree that age and gender are the two most important features, with gender being the top one. Meanwhile, the only important feature for AB is age, with positive importance.

AdaBoost has a low predicted probability branch formed by young people with cognitive difficulty and above-average school years (R1). Age and school years are the most important (and positive) features for those samples (R4). Interestingly, cognitive difficulty is not an important feature of this branch. All other models do not have a branch for those samples (R3). SVM and NN also have age and school years as the two most important (and positive) features, but they also consider cognitive difficulty as an important feature for those samples. RF and XG, on the other hand, assign negative importance to age and no importance to cognitive difficulty. The actual employment rate for those samples is 10\%, but AB classifies all of them as unemployed. This motivated a further investigation of how the models behave for people with cognitive difficulty. We found that all models have under 3\% False Positive Rate for those individuals, but they all have quite high False Negative Rates: around 95\% for LR, SVM, and AB; 80\% for NN and XG; and 74\% for RF. For reference, there were 2,745 samples with cognitive difficulty in the test set, with 14.5\% of them being employed. If we were indeed building models for decision-making (such as credit analysis), this find would indicate a case of model discrimination, sparking a further development of fairness-aware models for the cognitive difficulty characteristic. AB’s mapper also has two high-probability branches, but they have only under 15 samples each.

\subsection{Expert Feedback}
 Our study involved in-depth interviews with three explainability experts (referred to as P1, P2, and P3) to gather insights into our system's usability and potential in the context of their expertise. P1 and P2 have been industry Machine Learning practitioners for more than 5 years each and have considerable expertise in XAI. P3 is an academic with more than 3 years of research experience in the field of explainability. 
 Our interviews consisted of three distinct parts. First, we introduced our work along with its motivations and goals. Secondly, we conducted a guided demonstration of Case Study 1, as outlined in the preceding section. Throughout this demonstration, experts were encouraged to interrupt with questions, comments, and guidance as they saw fit. Finally, in the third part, we gathered the experts' perspectives on the tool's usability, utility, strengths, and limitations. The list of questions we asked in this part can be found in Appendix A.
 
\myparagraph{Design and Framework}
We received predominantly positive responses from the experts, who, while not visualization specialists, are well-versed in ML and XAI and are part of the tool's potential user group. P3 noted that the distance matrix allowed for a quick and comprehensive overview of the groups of explanations that are in agreement or disagreement (\textbf{R3}). Furthermore, P3 praised the feature that enables color coding of nodes in the mapper graph based on feature values (\textbf{R1}), as well as the ability to highlight specific nodes of interest and explore them in other linked views (\textbf{R2}). Despite the complexity associated with TDA, the experts affirmed that the visual encodings and provided interactions were ``user-friendly'' and ``intuitive'',  expressing that \system ``seamlessly facilitates comparison between explanations''. They also appreciated the tool's compatibility with Jupyter notebooks, making it easier to integrate into machine learning workflows (\textbf{R5}). Their perspectives as prospective users in our target domain helped validate the utility and design choices within \system. 
\vspace{0.01pt}

\myparagraph{Insights}
P1 appreciated the disagreement details between explanations our tool can communicate. Reflecting on Case Study 1, P1 stated: \emph{``You can see how some explanation methods are capable of detecting the importance of certain features that other methods cannot. And that we can find out which features those are and for which value ranges they matter''} (\textbf{R4}). Referring to the results shown in Figure \ref{fig:casestudy1-1} and Figure \ref{fig:casestudy1-2}, P3 stated \emph{`` MSinceMostRecentInqexcl7days seems to be the most important feature. I think the most important part is that we can also see the samples for which the methods disagree and investigate those samples''} (\textbf{R4}). Furthermore, P3 opined that without \system, it would require significant time and manual experimentation to arrive at such insights.

\myparagraph{Recommendations} P2 predominantly works with SHAP, expressing that the direct comparison with other explanation methods holds limited utility in their routine tasks. However, P2 acknowledged the potential value of the tool for fellow practitioners, proposing to augment its capabilities through the integration of SHAP dependency plots. P3's expertise lies in explainability for sequential data and expressed great interest in adapting the current framework to work with sequential or time series data. Additionally, P3 suggested the addition of an overview with small multiples of all mapper graphs.

Overall, the experts' feedback underscored the utility of our tool while also pointing to specific enhancements that could further broaden its applicability across various industry and academic contexts.

\section{Conclusions and Future Work} \label{sec:conclusion}

In this work, we presented \system, a topology-driven visual analytics tool to compare and analyze black-box ML model explanations. \system was developed in collaboration with industry practitioners and can be easily deployed within Jupyter Notebooks. Using intuitive interactions and visualizations, \system links the topological graphs of explanations back to the original dataset, the feature attributions, and the model predictions. Using a real-world credit dataset, we showcased how \system can be used to compare and evaluate black-box ML explanations. Furthermore, we illustrated how \system can also be used to compare and understand different machine learning models. Finally, we conducted evaluations with ML experts to help assess the utility of our work and understand avenues for future work. The code repository for \system can be found here: \url{https://github.com/PariSolunke/mountaineer}.

\myparagraph{Workflow} The interactions facilitated by our design gave rise to an iterative process of exploring the relationships between the mapper graphs, distance matrix, and data attributes. Experts found the back-and-forth workflow between these elements to be intuitive and insightful. Moreover, we showed this workflow's utility beyond just explanations in Case Study 2 - where we leveraged it to directly compare machine learning models' behaviors. Hence, the workflow we've outlined can be immensely valuable, not just during model validation but also in the post-deployment phase. It can offer ML practitioners a means to comprehend both explanation results and model behavior, aiding their decision-making process during model validation. Moreover, end-users can leverage the tool to delve into model and explanation behavior following deployment, empowering them to make informed choices regarding model selection. 

Our topology-centric workflow is versatile and can potentially generalize to many tasks involving the analysis of high-dimensional data. However, its effectiveness is contingent on the nature of the data involved as well as the choice of lens function. For example, in scenarios with high-dimensional but well-structured data, as demonstrated in Section \ref{sec:use-cases}, our workflow can effectively elucidate the underlying structure. In contrast, with sparse or highly noisy data, the topology may become less informative or computationally infeasible to generate. Additionally, our workflow cannot be applied to unstructured or semi-structured datasets. The utility of the generated topologies also heavily depends on the choice of the lens function. In the analysis of explanations for ML model predictions, the predicted probabilities are an obvious choice for the lens function, which might not be the case in other analysis tasks.

\myparagraph{Limitations} The limitations of \system include: 1) It only supports one vs one exploration of the topological representations. The analysis of multiple graphs simultaneously could be beneficial, as suggested by expert P3. 2) The visual scalability of the system is constrained to a range of only a few dozen explanation results. Consequently, there may be challenges related to scalability in situations where the evaluation of more than a few dozen explanations is required. However, it's worth noting that the majority of existing literature on assessing disagreement between explanations focuses on a limited number of explanations at a time \cite{krishna2022disagreement, pmlr-v196-xenopoulos22a, NEURIPS2022_22b11181, xenopoulos2022topological}. Furthermore, it's important to consider that generating explanations is a time and resource intensive task. Therefore, for most real-world use cases, this limitation should suffice in terms of the number of explanations to be evaluated. 3) Analyzing the feature importance and data distributions when there is a high number of features can be an arduous task. Currently, \system sorts the features based on their relevance in each view, which helps alleviate this problem. Still, this aspect could be improved. 4) Selecting the mapper parameters can be computationally costly, being the step with the longest runtime in the framework. This runtime scales with the number of samples, so \system can be impractical for larger datasets. Although 
our modified parameter selection strategy (Section \ref{sec:parameter-selection}) significantly decreases runtime in comparison with GALE (Appendix B), we believe that this process can be further enhanced to be more time efficient.

\myparagraph{Future Work} While we demonstrated the use of \system in the context of binary classification, it can also be employed for regression by using the normalized model predictions as the lens function. For future work, we intend to extend our system to multi-class classification problems. Furthermore, we intend to address the aforementioned visual limitations of our current system, making it easier for users to explore datasets with a high number of features, and also allow the user to compare multiple graphs at once as recommended by expert P3. Additionally, there have been numerous works\cite{motta2019hyperparameter, carriere2018statistical} addressing the problem of Mapper parameter selection; We believe we can further build upon our work and introduce a metric to quantify the quality of the Mapper parameter search as well as improve the runtime of the parameter search. Furthermore, we see potential in integrating visual guidance to enhance the parameter selection process. Finally, although we presented a comprehensive strategy for Mapper parameter tuning, the fine-tuning of hyperparameters for explanation methods remains unexplored in our current work, which represents an important direction for future investigation.

\section*{Acknowledgments}
This collaboration has been supported by a grant from Capital One. Silva’s research has also been supported by NASA; NSF awards CNS1229185, CCF1533564, CNS-1544753, CNS-1730396, CNS-1828576, CNS-1626098; and DARPA PTG and D3M. Nonato’s research has been supported by Sao Paulo Research Foundation (FAPESP)-Brazil (grant 2013/ 07375-0) and CNPq-Brazil (grant 307184/2021-8). Any opinions, findings, and conclusions or recommendations expressed in this material are those of the authors and do not necessarily reflect the views of DARPA, NSF, NASA, FAPESP, CNPq, or Capital One.

\bibliographystyle{IEEEtran}
\bibliography{IEEEabrv,mountaineer}



\mbox{~}

\onecolumn
\pagenumbering{gobble} 

\begin{appendices}

\section{Expert Feedback Questions}
Following are the questions we asked in the third part of our expert interview to gain their perspectives regarding our framework:
  \begin{enumerate}
     \item Do the views and interactions together facilitate the comparison of different explanations? 
     \item Are the interactions provided intuitive and easy to use?	
     \item Do you think the visualizations used gave you insights regarding the explanations?
     \item What were the major insights you gained from the demonstrated case study?
     \item Do you think these insights would be harder to get without a framework like Mountaineer?
     \item Can mountaineer be integrated into your daily workflow? 
    \item Do you have any other suggestions or recommendations for improvements to the tool? 
    
 \end{enumerate}

\newpage
\section{Scalability: Computing mapper parameters}
\setcounter{table}{0}
The parameter selection strategy first selects the Delta parameter (the distance threshold for the agglomerative clustering). Then, a Bootstrap is performed to select the resolution parameter.

Table~\ref{tab:parameter_selection_time:adult} shows the execution time of each step of the parameter selection and the computation of the mapper representation for Case Study 1. We also contrast the execution time for the parameter selection in Case Study 1 with GALE in the same computing environment. Table~\ref{tab:parameter_selection_time:acsemployment} shows the parameter computing time of our parameter selection strategy for Case Study 2. We were not able to test the parameter selection strategy used in GALE for Case Study 2, which uses a significantly larger dataset, due to insufficient computing resources.

The HELOC dataset used for Case Study 1 contains information about 9,871 credit applications with 24 continuous features. The ACSEmployment dataset used for Case Study 2 contains information on 159,299 adults residing in the state of New York with 2 continuous and 10 categorical features. 

For case study 1, our parameter selection strategy performed between 18-100 times faster for the various explanation results when compared to the parameter selection strategy presented in GALE\cite{pmlr-v196-xenopoulos22a}. Additionally, the runtime for our technique remains relatively stable when compared to the runtimes using GALE which exhibits many fluctuations.
\vspace{30pt}
\begin{table}[h]
    \renewcommand{\thetable}{A.\arabic{table}}

    \centering        
    \large

    \begin{tabular}{l|ccc|c}
        \multirow{2}{*}{Explanation Method} & \multicolumn{4}{c}{Execution Time (s)} \\
                             & Delta & Bootstrap & Mapper & GALE\\
        \hline
        Vanilla Gradient     & 0.66  & 13.23 & 0.03 & 263.57\\
        Gradient x Input     & 0.39  & 12.11 & 0.02 & 387.39\\
        Occlusion            & 0.52  & 12.21 & 0.02 & 794.09\\
        Guided Backprop      & 0.46  & 12.85 & 0.02 & 236.69\\
        LIME                 & 0.44  & 13.32 & 0.02 & 1308.51\\
        KernelSHAP           & 0.37  & 12.56 & 0.02 & 1448.22\\
        SmoothGrad           & 0.77  & 13.78 & 0.02 & 257.46\\
        Integrated Gradients & 0.39  & 12.95 & 0.03 & 553.41\\
    \end{tabular}
    \vspace{10pt}

    \caption{Execution time (in seconds) for computing the mapper parameters (columns Delta and Bootstrap) and the mapper representation (column Mapper) in Case Study 1. For comparison, we report the execution time for computing the mapper parameters using GALE\cite{pmlr-v196-xenopoulos22a}.}
    \label{tab:parameter_selection_time:adult}
\end{table}
\vspace{30pt}

\begin{table}[h]
    \renewcommand{\thetable}{A.\arabic{table}}

    \centering
    \large
    \begin{tabular}{l|ccc}
        \multirow{2}{*}{Model} & \multicolumn{3}{c}{Execution Time (s)} \\
                            & Delta & Bootstrap & Mapper\\
        \hline
        RBF SVM             & 29.85 & 3721.82 & 5.96\\
        Random Forest       & 28.37 & 1215.55 & 4.46\\
        Neural Net          & 39.06 & 1122.23 & 5.54\\
        AdaBoost            & 34.09 & 1328.09 & 5.29\\
        XGBoost             & 30.96 & 1268.44 & 5.76\\
        Logistic Regression &   -   &    -    & 3.88
    \end{tabular}
        \vspace{10pt}
    \caption{Execution time for computing the mapper parameters (columns Delta and Bootstrap) and the mapper representation (column Mapper) in Case Study 2.}
    \label{tab:parameter_selection_time:acsemployment}
\end{table}

\newpage

\end{appendices}

\vfill

\end{document}